\documentstyle[aaai]{article}
\begin{document}

\title{On the use of expectations for detecting and repairing 
human-machine miscommunication}

\author{ Morena Danieli \\
CSELT \\
Centro Studi E Laboratori Telecomunicazioni S.p.A.\\
Via G. Reiss Romoli 274 \\
I-10148 Torino, Italy \\
E-Mail: Morena.Danieli@cselt.it}

\date{}

\maketitle
\begin{abstract}
\begin{quote}
In this paper I describe how  miscommunication problems 
are dealt with in the spoken language system DIALOGOS.
The dialogue module of the system exploits dialogic expectations in a
twofold 
way: to model what future user utterance
might be about (predictions), and to account
how the user's next utterance may be related to previous ones in the 
ongoing interaction (pragmatic-based expectations). The analysis
starts from the hypothesis that the occurrence of miscommunication
is concomitant with two pragmatic phenomena: the deviation of the user
from the expected behaviour and the generation of a conversational
implicature.
A preliminary evaluation of a large amount of interactions between
 subjects 
and DIALOGOS shows that the system performance is enhanced by
the uses of  both predictions and pragmatic-based expectations.
\end{quote}
\end{abstract}

\section{The problem}

During the last few years it has been emerging that the success of
 spoken 
language systems is greatly improved by the contextual reasoning of
 dialogue modules.
 This tenet has spread through both the speech
  and the dialogue communities. Dialogue systems devoted to
spoken language applications are able to detect partial communication
 breakdowns by
other system modules, and that increases the robustness of
 human-machine
interactions by speech.

During oral interactions with computers, communication problems often
 arise 
after the occurrence of errors during the recognition  phase. Sometimes
these errors cannot be solved by the semantic module: the utterances
containing them are interpreted by the semantic analyser, but with an
information content different from 
the speaker's intended meaning. Detecting such miscommunications and
repairing them through initialization of appropriate repair 
subdialogues
is essential for the interaction to be successful. 

Most of the research in this area has been devoted to providing
the recognition and understanding modules with information
generated on the basis of the dialogue context. They predict what
the next user's utterance will probably be about: throughout this paper
I will refer to them with the name $''$predictions$''$. Sometimes they
 are
passed down to the acoustic recognition level in order to decrease the
huge number of lexical choices, sometimes they are used to help
in deciding on multiple semantic interpretations. More often they are
 used
during the contextual interpretation phase to accept or reject parser
output.

Although predictions have proved useful, they only grasp one side of
the miscommunication problem. Actually they are a means for
reducing recognition errors, and their use allows the avoidance of
 one of the
 potential sources of miscommunication. However during 
spoken human-computer interactions, the detection of miscommunications
may be outside the capabilities of the dialogue system, even though it
 uses
predictions; on the contrary, the user is usually able to detect any
 speech errors.
For example, in travel inquiry applications
words belonging to the same class, such as proper names of place, may
 be
highly confusable. When the dialogue prediction says that next user's 
utterance is likely to be about a departure place, this does not
 exclude
that the recognizer substitutes the actually uttered name with a
 phonetically
similar one. Only the user is able to detect such kinds of errors.
In
this situation the dialogue system should identify the user's
 detection of
miscommunication and
provide appropriate repairs. 

In this paper I will argue that the dialogue
module ability to detect user-initiated repairs is improved if the
 system
is able to capture the pragmatic
 phenomena that accompany user's detection of miscommunication. 
The paper offers an analysis of the pragmatic phenomena that occur
when users detect miscommunication
during task-oriented
 human-machine spoken dialogues.
The account exploits both predictions
and another notion of expectation that comes from the
cognitive-based research area. This notion refers to conversants'
 beliefs about
the relation of future utterances with previous ones in a dialogue.
 A computational
interpretation of this notion has been done in the model proposed
by \cite{Roy95}, where it is suggested that  speech community
predictions and  cognitive based expectations are complementary
notions. This paper claims that accounting
for these two notions is useful for detecting and solving actual 
breakdowns in
user-system communication. The working hypothesis is that in
 task-oriented dialogues
 miscommunication  often generates
conversational implicatures. I will show how they are dealt with by
 the dialogue
 module of DIALOGOS, 
an information inquiry spoken language system implemented by Cselt
speech recognition and understanding group. By using the telephone, 
the system may be
 used to
access the data base of the Italian public railway
company. I will report dialogue examples and experimental data that
show the effectiveness of the proposed analysis in task-oriented
human-machine spoken dialogues.

\section{The working hypothesis}

The 
conceptual background of this approach is inspired by the Gricean
 principles of 
conversation \cite{LC}. The user modelling of the dialogue module 
 of DIALOGOS
 is based on the assumption that
both the system and the user are active agents of the communicative
 process;
in particular, it is assumed that both of them observe the Cooperation
 Principle
(CP)
in order to achieve the general goal of their linguistic interchange,
 i.e.
to access a database to get the information that the user needs.

The system predictions
are modelled on the basis of the CP. At each stage of the dialogue
 with 
the user the system expects that user's reaction conforms with three
of the original Gricean conversational
maxims, re-interpreted in the context of human-machine communication in
 \cite{Norman}. For example, in this application domain,
throughout  the dialogue the system expects that:

\begin{itemize}
\item each user turn is not over or under-informative,

\item each user turn communicates user's $''$true'$''$ needs (i.e. to
 receive
 train timetables from place X to place Y, if the two places have been
 confirmed), 

\item is pertinent to the focus in hand. 

\end{itemize}

However, as remarked earlier, in oral 
human-machine dialogue the communication process
 may be disturbed
by several factors, the most usual of ones are recognition errors. 
The user's detection of breakdowns and errors of the system has precise
empirical consequences: these concern both user's behaviour and her
cognitive demand on the continuation of the interaction. In particular,
these empirical consequences may be summarized as follows:

\begin{itemize}

\item the user utterance does not match dialogue predictions;

\item the user asks the system to come back to the 
interpretation context where miscommunication occurred;

\item both the user and the system should engage in a clarification
      subdialogue before continuing their interaction. 

\end{itemize}

 Since dialogue predictions are generated by taking into account the
respect of the above listed maxims, and given that the system assumes
 that
 the user continues
to respect the CP, each deviation of the user's behaviour from dialogue
predictions is ideally interpreted as a signal of the potential
 occurrence
of miscommunication. By co-operating in the achievement of the goal of
 the 
conversation, the user intentionally violates the maxims when she
 detects 
that a misunderstanding occurred in previous turns. Accordingly to the
 conceptual
background here adopted, the intentional violation of the
 conversational 
maxims gives place to a conversational implicature. 

The working
hypothesis is that the occurrence of a miscommunication goes along
 with two
 pragmatic counterparts: the first one is
 the deviation of the user's behaviour from the dialogue predictions,
 the other one
 is the generation of a conversational
 implicature.
the turn 
 For being $''$conversational$''$ an implicature has to be inferable
on the basis of the contextual knowledge. I suggest that in order to
 recognize
 conversational 
implicatures, the dialogue system should embody a model of user 
expectations that gives insights on how to relate user correction
turns to the previous (normal) turns in an ongoing dialogue.

\section{Examples of miscommunication}

\subsection{Non-understanding}

The kind of miscommunications that occur during oral interactions of 
subjects with spoken language systems is usually caused by an actual
breakdown in the flow of the dialogue. The most usual
 miscommunications may be
labelled as $''$misunderstanding$''$ and $''$non-understanding$''$. 
 The latter are
usually
less problematic.

\begin{figure}[htb]
\begin{small}
\begin{tabular}{l} \hline \\

{\bf T1 S}: Automatic Railway Information System. Please\\
	    speak after the tone. Please give your point of\\
	    departure and your destination. \\

{\bf T2 U}: Then ... What time .. I mean from [NOISE] eight\\

            \it{$<$Recognition and understanding failure$>$}\\

{\bf T3 S}: Sorry, I did not understand.  Please give your \\
	    point of departure and your destination. \\

{\bf T4 U}: I want to leave from Milano in the evening.\\
            I 'd like to know departure [NOISE] from\\
	    Milano to Roma.\\
            \it{$<$confirmation=NO, departure-city=MILANO,} \\
            \it{ arrival-city=ROMA, departure-time=EVENING$>$}\\ 

{\bf T5 S}: Do you want to travel from Milano to Roma \\
	    leaving in the evening? \\

{\bf T6 U}: Exactly. Around eight p.m. \\

           \it{ $<$confirmation=YES hour=EIGHT$>$ } \\

{\bf T7 S}: Intercity 243 leaves from Milano Centrale at \\
	    20 past 8 p.m.; it arrive at Roma Termini 6 a.m.\\
\\ \hline
\end{tabular}
\end{small}
\caption{\label{D1ex}Example of non-understanding}
\end{figure}

In the dialogue reported in Figure ~\ref{D1ex} non-understanding
 occurred in the 
turn T2:
due to environment noise,
the fragmentary utterance 
was not understood by the recognizer and no semantic representation
was sent to the dialogue module by the parser. This symptom of actual
 breakdown 
causes the system to trigger an informative speech act of
 non-understanding 
and a requestive speech act for obtaining departure and arrival place
 names (T3).

In non-understanding
the system completely fails to interpret a user's utterance; that may
happen either because a speech recognition error occurred or because 
the linguistic
processor was not able to interpret it, for example if the semantic
content of the sentence was out of its semantic coverage. 
Non-understanding is
usually recognized by the dialogue system as soon as it happens.
 Within speech
understanding systems 
 non-understanding usually gives
place to second-turn repairs \cite{Schegloff92}. For example, in the
 DIALOGOS
system the dialogue module realizes that a non-understanding has
occurred because it fails to receive any semantic interpretation for
 the 
user's sentence uttered immediately before. In this case it 
enters a repair action by informing the user that her utterance was not
understood and asking her to rephrase the sentence. At
present the dialogue module does not know if the non-understanding
 was caused 
by a failure of the speech recognizer (because of out-of-vocabulary
 words,
mispronouncing, unhearings, and so on) or by a failure of the parsing 
module. As a consequence, the user is addressed with a generic
 information
of non-understanding.

\subsection{Misunderstanding}

Some kinds of failures in recognition and understanding cause actual
 human-machine
misunderstanding. Let us consider the dialogue in  Figure ~\ref{D2ex}. 
The user utterance T2 has been recognized as  $``$Arona$''$  instead of
 $``$Roma$''$ (in Italian they may be confusable). This causes a 
third-turn repair.
 In the turn T4 the user, instead of giving the
departure place name, realizes that the sentence uttered in the turn 
T2 was not
 correctly recognized and
she repeats the arrival city. In producing T3, the system generated a
 prediction
about the departure city, but user's turn T4
causes it to recover. This is done by reinterpreting T4 within the
 interpretation
context initially generated for T2. Moreover 
the expectations generated in interpreting T4 are different from the
ones generated for T2. In T4 the system realizes that a
 misunderstanding   
happened: it discards predictions about obtaining the
departure city by refocusing the interpretation
context on the possible confirmation of the arrival city.

\begin{figure}[htb]
\begin{small}
\begin{tabular}{l} \hline \\

{\bf T1 S}: Automatic Railway Information System. Please\\
	    speak after the tone. Please give your point of\\
	    departure and your destination. \\

{\bf T2 U}: I want to go to Roma. \\

    \it{$<$ *arrival-city=ARONA $>$} \\

{\bf T3 S}: To Arona. What is your point of departure? \\

{\bf T4 U}: I said that I was going to Roma. \\
    \it{$<$ arrival-city=ROMA $>$} \\

{\bf T5 S}: Are you going to Roma?\\ 
\\ \hline
\end{tabular}
\end{small}
\caption{\label{D2ex}Example of misunderstanding due to substitution}
\end{figure}

\section{Identifying misunderstanding}

The example reported in Figure ~\ref{D2ex} is an actual piece of
 dialogue 
(translated into English)
between a subject and the DIALOGOS system. Let us now describe how
the system deals with the misunderstanding. In the turn T3 the
dialogue system generates a so-called 
$''$confirmation plus initiative$''$
speech act. This means that in a single dialogue
turn the system tries to get both a confirmation of what it understood 
from the previous user turn, and other piece of information it needs
 to access the
database. The selection of this speech act generates
a large number of expectations. The most interesting for the present
 discussion
are listed below. For brevity, in what follows I have
 considered only the predictions of statements with more informational
content than simple confirmation and denial by the adverbs $''$yes$''$ 
and $''$no$''$.

\begin{enumerate}

\item A statement about the name of the departure city;

\item A statement about the name of the departure city and other
 required parameters, such as  the date and the time of departure;

\item An explicit confirmation of the arrival, and the departure city;

\item An explicit denial of the understood arrival city
 and a request for another arrival city;

\item An explicit denial of the understood arrival city;

\item A statement including a new arrival city plus the departure city;

\item A statement including a new arrival city;

\end{enumerate}

As we can observe, the expectations from (1) to (7) may be related to
predictable lexical counterparts. In other words, they may be sent to
 the
recognition level in the form of word class predictions, in order to
 be used for
constraining the search space of the speech recognizer.

In addition, we can observe that at the dialogue level the realization
of expectations (6) and (7) implies something more
than the prediction of a lexical class.  The linguistic realizations
 of (6) and
(7) have the form of affirmative statements and the illocutionary 
force of
disconfirmations.  In both cases a conversational implicature is 
generated. 
The illocutionary goal of statements realizing (6) and (7) 
imply a precise cognitive demand by the user.
Let us again consider Figure ~\ref{D2ex}. T4 realizes dialogue
 expectation (7):
the user asks the system
to reinterpret T4 in the
context of T2. Notice also that the user's beliefs about the
 continuation
of the interaction are slightly different for (6) and (7): in the
 case of 
(6) the focus of attention of both user's and system  are still on
 the same
speech act, i.e. the disconfirmation of the arrival city and the
 proposal of a
departure city. On the contrary, in (7) the user expects that
 {\it before} 
requesting a new parameter
the system will be able to correct its previous  misunderstanding.
In other words, she demands a focus restriction in order to correct
 the
misunderstanding. An appropriate  system answer  in situations like
 (6) and (7) requires that
the system be able to grasp the illocutionary force of user statements
 in
order to redirect the dialogue strategy. 

\section{Dealing with misunderstanding}
 
Let us consider in Figure ~\ref{D3ex} another excerpt from an
 interaction
 between  DIALOGOS and  a subject.

\begin{figure}[htb]
\begin{small}
\begin{tabular}{l} \hline \\

{\bf T1 S}: Automatic Railway Information System. Please\\
	    speak after the tone. Please give your point of\\
	    departure and your destination. \\
 
{\bf T2 U}: From Milano to Roma. \\

   \it{ $<$ departure-city=MILANO *arrival-city=ARONA $>$} \\

{\bf T3 S}: Do you want to travel from Milano to Arona? \\

{\bf T4 U}: From Milano to Roma. \\
    \it{$<$departure-city=Milano arrival-city=ROMA $>$} \\

{\bf T5 S}: Do you want to arrive in Roma?\\ 
\\ \hline
\end{tabular}
\end{small}
\caption{\label{D3ex}Example of repair with repetition}
\end{figure}

In the turn T3, DIALOGOS prompts the user for confirmation  with a 
Yes$/$No
 question.
At this point the system linguistic predictions concern a confirmation
 or
 a negation
speech act by the user. In T4 the user produces an affirmative 
sentence which has
the illocutionary force of a disconfirmation. By 
analysing a large number of interactions between users and the
 DIALOGOS 
system, we actually realized that in this application context Italian 
speakers very often start a repair without an explicit negation word 
such as $``$No$''$: on the contrary, they repeat their previous
 utterances louder.
Notice that in Italian
repetitions mostly work for confirmation \cite{BAZZA}, but in that
case the intonational contour is unstressed.

As proposed above, in order to recognize the dialogic repetition of
 the turn T4
as a user-initiated repair, the dialogue system has to be able to take
 into 
account both the deviation of the user from the expected behaviour,
and the illocutionary force of her utterance (i.e. the denial of the
 system
interpretation of T2). This deviation is the signal of an actual
 misunderstanding
and it is related to the presence of a conversational implicature.

The dialogue 
system initially interprets T4 with respect to its set of expectations.
As soon as it realizes that the matched expectation realizes an
 implicature,
it reinterprets T4 within the context of the utterance whose
 interpretation is
 now denied, i.e. the system traces back to the 
interpretation context of T2. The fact that T4 is now being
reinterpreted in the context of T2 causes a refocusing of the
 dialogue context.
In the present DIALOGOS version the refocusing mechanism is allowed
 through
 the maintenance of both a linguistic history and a context hierarchy
 (further
details of this are reported in the section that describes the 
dialogue system).

All the examples discussed above concern third-turn repairs; however, 
repairs from misunderstanding may 
occur also in subsequent turns, as the example in 
Figure ~\ref{D4ex} shows.

\begin{figure}[htb]
\begin{small}
\begin{tabular}{l} \hline \\

{\bf T1 S}: Automatic Railway Information System. Please \\
	    speak after the tone. Tell me from where you \\
	    want to leave and where you want to go.\\

{\bf T2 U}:... leaving from Pisa at eight [NOISE] this \\
	    morning [NOISE] Firen-\\

        \it{ $<$ *departure-city=PISA-AEROPORTO} \\
	 \it{  departure-time=MORNING date=TODAY $>$} \\

{\bf T3 S}: From Pisa Aeroporto. Where are you going to?\\

{\bf T4 U}: To Firenze. \\
    $<$arrival-city=FIRENZE$>$ \\

{\bf T5 S}: From Pisa Aeroporto to Firenze?\\

{\bf T6 U}: From Pisa ... Stazione Centrale to Firenze \\
       \it{ $<$departure-city=PISA dep-station=CENTRALE } \\
	 \it{  arrival-city=FIRENZE$>$} \\

{\bf T7 S}: From Pisa Centrale?\\

{\bf T8 U}: Yes\\
 \it{$<$ confirmation=YES $>$ }\\ 
\\ \hline
\end{tabular}
\end{small}
\caption{\label{D4ex}Example of misunderstanding}
\end{figure}

In this example the sentence uttered in the turn T2 was badly
 recognized. For the
sake of simplicity, let us concentrate on what happened about the
 names of the
departure and arrival city. Due to disturbances over the phone line, 
the uttered arrival city name ($''$Firenze$''$) 
was not decoded. Moreover the phrase $''$from Pisa at eight$''$ (in
 Italian,
 $''$da Pisa alle otto$''$) was recognized as $''$Pisa Aeroporto$''$. 
The system expectations were not satisfied since the name of the
 arrival city
was not acquired. Then the dialogue system
decided to trigger a confirmation plus
initiative speech act in order to obtain the missing parameter
and to confirm departure. 
That choice resulted in the generation of T3. At this dialogue stage 
the 
set of expectations were the ones enumerated in the previous paragraph,
 although
applied to the arrival city parameter. In T4 the user
offered an arrival city. This matched the first expectation of our
 list. At this
stage neither the user nor the system had grasped the inconsistency
 concerning
departure. Since it had not obtained explicit confirmation for 
departure,
the  dialogue system addressed the user with the Yes$/$No question of
 turn T5.
The contextual interpretation of the
user utterance T6 detects the explicit confirmation of  the arrival
 city and
the misunderstanding that had occurred during the recognition of the
 departure city.  
The latter is refocused again
in T7, and  the user is addressed with a new  Yes$/$No question.

\section{Architecture of DIALOGOS}

DIALOGOS is a real-time task-oriented system composed of the following
 modules:
 the acoustical front-end, the linguistic processor,
 the dialogue manager and the message generator, and
 the text-to-speech synthesizer.  Its vocabulary size is about 3,500
 words 
including 2,983 place names.
The acoustical front-end and the synthesizer are connected to the
 telephone
network
through a telephone interface, while the dialogue manager is connected
 to a 
Computer Information System to obtain information about Italian Railway
time-tables. The acoustical front-end performs feature extraction
 and acoustic-phonetic decoding. The recognition module is based on a
 frame
 synchronous
Viterbi decoding,
where the acoustic matching is performed by a phonetic neural network.
During the recognition, it makes use of language models that are
 class-based
bigrams trained on 30K sentences. The training data were partially
 derived
from the trial of a previous spoken language system applied to the 
same domain
\cite{baggiacrim94}.

The linguistic processor starts from the best-decoded sequence,
and performs a multi-step robust partial parsing.
In this strategy, partial solutions are accepted according to
the linguistic knowledge \cite{ICASSP93}. 
At the end of the parsing stage a deep semantic 
representation for the user utterance is sent to the dialogue module,
 that
will be described in the next subsection.

The dialogue module performs contextual interpretation and generates 
the answer which is sent to the 
text-to-speech synthesizer Eloquens \cite{Quazza93}, that contains
 specific
 prosodic rules
for the Italian language.

\subsection{The dialogue system}

The dialogue strategy of DIALOGOS has been designed 
both to maintain the control of the cooperative interaction and
to leave the expert user the freedom to guide the interaction.
 Referring to 
the classification reported in \cite{Smith95}, the DIALOGOS system 
is able to support directive, suggestive and declarative
initiative modes. It only exploits the directive mode in order to deal 
with repetitive problems at the recognition and understanding levels. 
The directive
mode is implemented by automatic switching to the isolated word
 recognition
modality.

The flow of a typical interaction with the system
may be divided into two main stages: 
the acquisition of a set of parameters, used to
select a reasonably small set of objects of interest to the user, and 
the presentation of the information related to the
retrieved objects. These two stages may lead to the
combination of several subdialogues, each of them with its 
own purpose. The system is able to move properly
through nested subdialogues thanks to the maintenance of a dialogue
 history
which relates the focus of the current utterance
to the appropriate interpretation context. A detailed discussion of
the dialogue strategy is given in \cite{Bet93}.

The system performs
contextual interpretation on the basis of the model of user-system
 interaction
discussed above. The model has been 
implemented using a Transition Network formalism. Each node in the
 network
represents a state in the dialogue: associated actions are executed
 when the
interaction comes to that state. The actions are declared in a 
library of
functions written in C. There are specific actions for each dialogue
 functionality,
including contextual interpretation, updating of the context
 hierarchy (see below),
and generation
of dialogue predictions and expectations. A state transition is
 executed
when the conditions
associated with the arc are satisfied. Conditions are also library
 functions: they
check both the current state of the dialogue model and the content of
 the
user's utterance, in order to direct the interaction into a new
 coherent state.
Default transitions are associated with each node.

The contextual interpretation of user utterances takes into account the
linguistic history and the global and local (active) focus
 \cite{Grosz}, \cite{McCoy}.
Each time a user utterance has to be interpreted, all the information
useful for its interpretation (specifically, its semantic content and
some surface information that may be used to solve references) are 
stored in a cycle-structure \cite{BaggiaGerb}. At each point of
the dialogue, the linguistic history consists in the history of the
previous cycle-structures. The interpretation of the utterance causes
the creation of a local focus structure which is linked to the
 cycle-structure
that has caused it \cite{Bet93}. The focus structures are
 hierarchically 
organized in a tree (that we name $''$context hierarchy$''$),
whose root represents the global focus at the beginning
of the dialogue. A new node in the tree, that is a new active focus,
 is created and
linked in the hierarchy when the user operates a focus restriction or
 a focus
shifting. The correct interpretation context of an utterance can be the
active focus, if the utterance refers to the objects currently focused;
when there is discrepancy between utterance focus and active focus, the
hierarchy is climbed up for checking the semantic and pragmatic 
consistency between the current utterance and the previous ones.
The first node where the consistency is verified is chosen as the 
utterance interpretation context.

\section{Experimental Data}

 Recently DIALOGOS has been tested in a large field trial. Five 
hundred 
subjects
of different ages and levels of education
called the system from all over Italy. They had never used
a computerized telephone service before. Each of them made 
three calls to the system. The experimental material has been
 transcribed and
it is currently being evaluated. 

Subjects called the system from their own places (80\%), 
from public telephone booths (10\%), 
and by mobile phones used in noise environment (such as street,
underground stations, and so on). They received a single
page of printed
instructions which contained a brief explanation of the service 
capabilities.

Each subject had to plan a trip from a city A to a city B in
 a certain 
date and time: they had
to find out departure and arrival times of trains that satisfy their
 needs. 
The departure and the destination were specified in the scenarios, on
 the
contrary the subjects had the freedom to choose the date and
 the time of the travel.

Some features of the 
dialogues collected in the test and already evaluated are
shown in Table ~\ref{D5ex}. 
The total number of evaluated dialogues, the number of
continuous speech utterances, the number of isolated words,
and the average number of continuous speech
utterances per
dialogue are reported.

\begin{table}[htb]
\begin{center}
\begin{tabular}{|l|c|c|c|} \hline \hline
 No. of & No. of &  No. of & No. of \\ 
  Dialogs & Utterances & Iso. Words & Utt. per Dial.\\ \hline
  923 & 9,124 & 442 & 9.8 \\
\hline \hline
\end{tabular}
\end{center}
\caption{\label{D5ex}Corpus}
\end{table}

A user-system interaction was considered  successful
when the user received all the information she needed. In particular,
the transaction success is the measure of the success of the system
in providing users with the train timetables they required.
A definition of transaction success and related dialogue evaluation
metrics is given in \cite{Danieli95}.  The rate of 
transaction success for the 923 evaluated dialogues is 84\%. 
We consider this result a promising starting point for further work
within the conceptual framework described in this paper.

\section{Conclusions}

This paper discusses the use of expectations for dealing with
 miscommunication
in spoken dialogue
systems.  It has been suggested that the
deviation
of the user's behaviour from predictions, along with the generation of
a conversational implicature, are symptoms of actual miscommunication. 
The proposed analysis  exploits two 
 notions of expectations. The first one refers to predictions: they
have been largerly used by the speech community to reduce speech and
semantic interpretation errors. The second expectation notion gives
 insights on how
to relate future user utterance to previous ones. The described
 approach
 has been implemented in 
the dialogue module of the DIALOGOS spoken language system.

The level of appropriateness of the  system's answer when
miscommunication occurs is greatly enhanced when the spoken dialogue
 system 
is able to profit from both kinds of expectation.
Some experimental data support this
claim. Preliminary results
(based on 923 
dialogues with naive users) show that the rate of 
transaction success of the system is 84\%. 

Although this analysis is only a first step towards an adequate
 handling of
miscommunication within our automatic speech recognition system, it
 has been
shown to be useful
in improving the overall performance of the system. 

\section{Acknowledgements}
I would like to thank all the people that designed and implemented the
current version of the DIALOGOS system: Dario Albesano, Paolo Baggia,
 Roberto Gemello, and Claudio Rullent. A special thanks to Elisabetta 
Gerbino who discussed and implemented with the author the preliminary
version of the
dialogue system. Carla Bazzanella and Elisabeth Maier,
 who carefully commented 
on an earlier draft
 of this paper, are gratefully acknowledged.

\end{document}